\documentclass[runningheads]{llncs}
\usepackage{graphicx}
\usepackage{comment}
\usepackage{wrapfig}
\usepackage{amsmath,amssymb} 
\usepackage{color}

\usepackage{cite}
\usepackage{subfig}
\usepackage{url}
\usepackage{array}
\newcolumntype{P}[1]{>{\centering\arraybackslash}p{#1}}
\usepackage{multirow}

\begin{document}
\pagestyle{headings}
\mainmatter
\def\ECCVSubNumber{6994}  

\title{Representative-Discriminative Learning for Open-set Land Cover Classification of Satellite Imagery} 

\titlerunning{Representative-Discriminative Learning for Open-set Classification}
%
\author{Razieh Kaviani Baghbaderani\inst{1} \and
        Ying Qu\inst{1}\thanks{Corresponding author} \and
        Hairong Qi\inst{1} \and
        Craig Stutts\inst{2}}
\authorrunning{R. Kaviani Baghbaderani et al.}
%
\institute{The University of Tennessee, Knoxville, TN, USA\\
\email{\{rkavian1,yqu3,hqi\}@utk.edu}
\and
Applied Research Associates, Raleigh, NC, USA\\
\email{cstutts@ara.com}}
\maketitle

\begin{abstract}
Land cover classification of satellite imagery is an important step toward analyzing the Earth's surface. Existing models assume a closed-set setting where both the training and testing classes belong to the same label set. 
However, due to the unique characteristics of satellite imagery with extremely vast area of versatile cover materials, the training data are bound to be non-representative.  
In this paper, we study the problem of open-set land cover classification that identifies the samples belonging to unknown classes during testing, while maintaining performance on known classes.
Although inherently a classification problem, both representative and discriminative aspects of data need to be exploited in order to better distinguish unknown classes from known. 
We propose a representative-discriminative open-set recognition (RDOSR) framework, which 1) projects data from the raw image space to the embedding feature space that facilitates differentiating similar classes, and further 2) enhances both the representative and discriminative capacity through transformation to a so-called abundance space.
Experiments on multiple satellite benchmarks demonstrate effectiveness of the proposed method.  
We also show the generality of the proposed approach by achieving promising results on open-set classification tasks using RGB images. 
\end{abstract}

\section{Introduction}

Recent advancements in computer vision, especially the advent of Convolutional Neural Networks (CNN), have significantly improved the performance of image classification \cite{krizhevsky2012imagenet,  he2016deep, simonyan2014very}, detection \cite{redmon2016you, ren2015faster}, and segmentation \cite{long2015fully, chen2017deeplab} tasks, enabling their deployment in many different fields. One of such field of applications is satellite image analysis that includes resource management, urban development planning, and climate control.
Land cover classification or material classification is one of the building blocks of satellite image analysis, providing essential inputs to a series of subsequent tasks including object segmentation, 3D reconstruction and modeling, as well as texture mapping. Supervised land cover classification involves categorization of multispectral or hyperspectral image pixels into predefined material classes, e.g., asphalt, tree, concrete, water, metal, soil, etc.
Note that both multispectral and hyperspectral images (MSI and HSI) try to provide additional spectral information, beyond the visible spectra, to reveal extra details and compensate for the coarse  spatial resolution of these images. 

\begin{figure}[t]
\centering
 \includegraphics[scale=0.42]{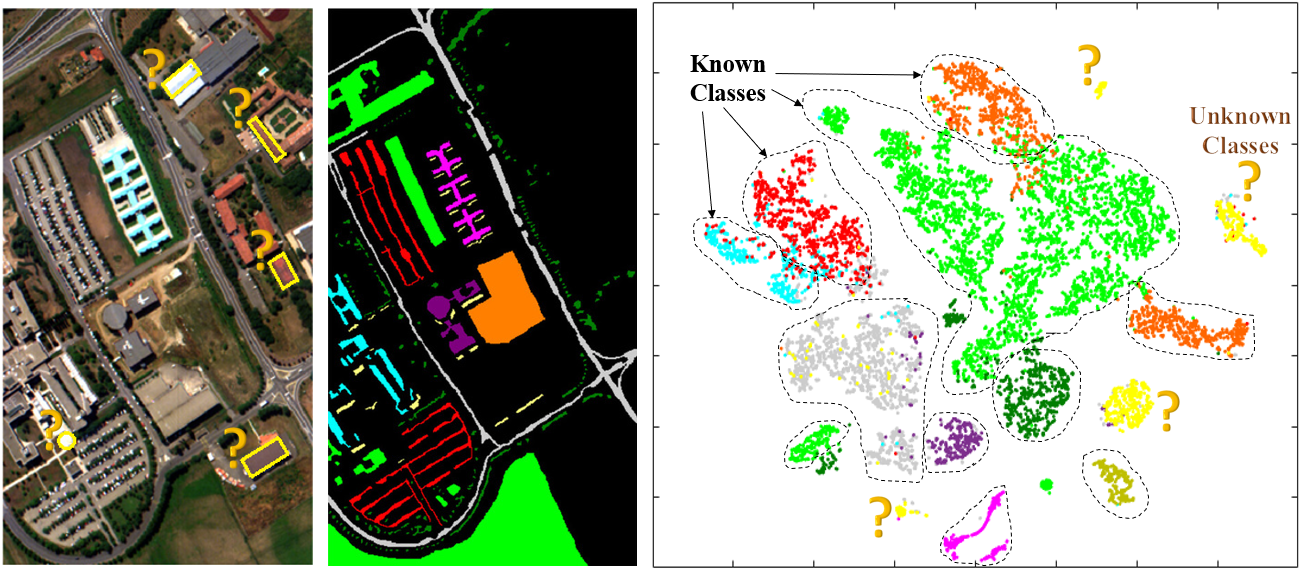}
 \caption{Open-set land-cover classification: Data samples corresponding to 
 ground truth categories are from the known class set (\textbf{K}). It is likely that some categories are not known during training and will be encountered at testing, \textit{i.e.}, samples from unknown class set (\textbf{U}). The goal is to identify pixels coming from (\textbf{U}), while correctly classify any pixel belonging to (\textbf{K}). From left to right, a satellite image  from the Pavia University dataset \cite{PUlink} showing unknown materials surfaces with yellow bounding boxes, the ground truth labels, and visualization of the feature space for both known and unknown classes using tSNE \cite{maaten2008visualizing}.} 
 \label{fig:motivation}
\end{figure}
 
Although inherently a classification problem,  material classification in satellite imagery faces a unique challenge: 
the vast area  covered by the satellite imagery makes the task of generating representative training samples almost impossible, as there are a large variety of materials existed on the Earth's  surface, especially those not well-exploited regions. 
Therefore, one of the most essential capabilities of land cover classification is to be able to automatically identify which test image and which area or pixel location of the image, has a higher probability of hosting new classes of materials.  This would provide essential guideline to human operators in collecting training samples for the new classes.

The vast majority of existing works for land cover classification have been done under the ``static closed world'' assumption, meaning that the training and testing sets are drawn from the same label set. As a result, a system observing any unknown class is forced to misclassify it as one of the known classes, thus, weakening the recognition performance. A more realistic scenario is to work in a non-stationary and open environment that not all categories are known \textit{a priori} and testing samples from unseen classes can emerge unexpectedly.  Recognition of known and unknown pixels in a given image and correctly classifying known pixels is defined as ``open-set land cover classification''. Fig. \ref{fig:motivation} explains this process using a real-world satellite image.

\begin{figure}[h]
\centering
\includegraphics[scale=0.5]{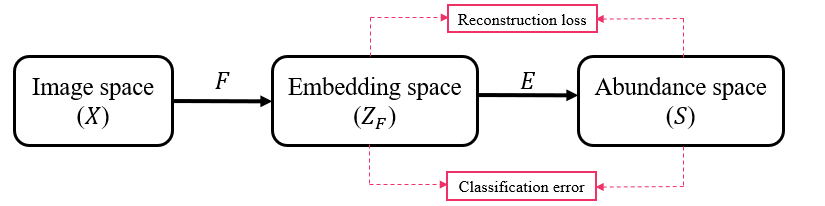}
\caption{Representative-discriminative learning through the transformation among 3 spaces: the raw image space, the embedding space, and the abundance space.}
\label{fig:space}
\end{figure}

In this paper, we present a multi-tasking representative-discriminative open-set recognition (RDOSR) framework to address the challenging land cover classification problem, where both the representative and discriminative aspects of data are exploited in order to best characterize the differences between known and unknown classes. We propose the representative and discriminative learning among three spaces, as shown in Fig.~\ref{fig:space}, including 1) the transformation from the raw image space to the embedding feature space,  
and 2) the transformation from the embedding feature space to a so-called abundance space.  
See Supplement A for an illustration of the effect in different spaces. 

The contributions of this paper are thus summarized as follows: First, unlike other open-set recognition methods applied on the raw image space directly, we propose to first learn a classification network that would transform from the raw image space to an embedding feature space such that a more discernible input is fed into the subsequent open-set learning network. Second, we propose to use the so-called Dirichlet-net to transform data from the embedding feature space to the abundance space. Due to the resolution issue, each pixel in a satellite image covers a large area with more than one constituent material, resulting in ``mixed pixel''. The mixtures are generally assumed to be a linear combination of a few spectral bases, with the corresponding mixing coefficients (or abundances). This way, instead of looking at the mixed pixel, we study the mixing coefficients of each spectral basis in making up the mixture. Thus the abundance space provides a finer-scale representation. 
Third, to the best of our knowledge, this work is the first attempt to address the critical open-set land cover classification problem essential for analyzing the Earth's surface. Fourth, while the proposed method was motivated by satellite imagery analysis, it is generalizable to RGB images and achieves promising results.
\section{Related Work}

\textbf{Conventional Land Cover Classification.} 
These methods mainly employ a traditional classifier on spectral information, with its discriminative power further enhanced through feature engineering algorithms such as minimum noise fraction (MNF) \cite{green1988transformation}, independent component analysis (ICA) \cite{comon1994independent}, morphological profiles \cite{plaza2005dimensionality}, and spectral unmixing \cite{luo2009unsupervised, dopido2011unmixing, baghbaderani2019hybrid}. The advent of deep learning has enabled the extraction of  hierarchical features automatically and achieved unprecedented performance. \cite{hu2015deep,song2019land} applied a 1D-CNN framework in the spectral domain to take into account the correlation between adjacent spectral bands. Several works use a patch surrounding the desired pixels by adopting a 2D-CNN structure \cite{cao2018hyperspectral, sharma2016hyperspectral} to incorporate the spatial correlation as well. More recently, integration of both spectral and spatial domains using 3D-CNN structures have been employed to further improve the classification accuracy \cite{hamida20183, zhong2017spectral}.

Although each approach has its own merit, all the existing land cover classification approaches work under the closed-set assumption where the training and testing sets share the same label set.

\noindent\textbf{Open-set Recognition.} 
Open-set recognition has gained considerable attention due to its handling of unknown class samples based on incomplete knowledge of the data during model training. 
Early studies are based on traditional classification models including Nearest Neighbor, Support Vector Machine (SVMs), Sparse Representation, etc. The open-set version of Nearest Neighbor was developed based on the distance of the testing samples to the known samples \cite{junior2017nearest}. The SVM-based approaches employed different regularization terms or kernels to detect unknown samples \cite{scheirer2014probability,junior2016specialized}.
In \cite{zhang2016sparse}, the residuals from the Sparse Representation-based Classification (SRC) algorithm were used as the score for unknown class detection.

In the context of deep networks, \cite{bendale2016towards} employed a statistical model to calibrate the SoftMax scores and produced an alternative layer, called OpenMax.
\cite{hassen2018learning} improved upon the OpenMax layer approach by maximizing the inter-class distance and minimizing the intraclass distance on the penultimate layer. 
The work of \cite{shu2017doc} proposed a k-sigmoid activation-based loss function for training a neural network to be able to find an operating threshold on the final activation layer.
\cite{yoshihashi2019classification} incorporated the latent representation for reconstruction along with the discriminative features obtained from a classification model to enhance the feature vector used for open-set detection.
Unlike previous methods, \cite{oza2019deep} utilized reconstruction error obtained from a multi-task learning framework as a detection score. Recently, \cite{perera2020generative} proposed using self-supervision and augmented the input image to learn richer features to improve separation between classes.

More recent works try to simulate open-set classes in order to provide explicit probability estimation over unknown classes. Ge et al. \cite{ge2017generative} extended OpenMax \cite{bendale2016towards} by synthesizing the unknown samples using a Generative Adversarial Network (GAN) based framework. Along the same line, \cite{neal2018open} proposed the counterfactual image generation (OSRCI) framework which employs a GAN to generate samples placing between decision boundaries that can be treated as unknown examples.
\cite{oza2019c2ae} proposed class-conditioned auto-encoder (C2AE) algorithm where conditional reconstruction helps learning of both known and unknown score distributions.

It should be noted that there are related problems in the literature including outlier detection \cite{xia2015learning, you2017provable} and anomaly detection \cite{chalapathy2017robust, golan2018deep} that have some overlap with open-set recognition and can be treated as a relaxed version of open-set recognition. These problems assume the availability of one abnormal class during training. However, general open-set recognition problems usually do not provide information about the type or the number of the unknown classes in advance.

\section{Proposed Approach}
We propose a representative-discriminative open-set recognition (RDOSR) structure, as shown in Fig.~\ref{fig:framework}. The network mainly consists of two components, 1) a closed-set embedding component to project the data from the original image domain to the embedding domain, such that different classes with similar spectral characteristics are more distinguishable, and 2) a multi-task representative-discriminative learning component to  learn a better representation scheme at a finer scale in the abundance space, such that unknown classes can be better differentiated from  known classes.

\subsection{Network Architecture}
\label{sec:network}

\begin{figure}[!t]
\centering
\includegraphics[scale=0.5]{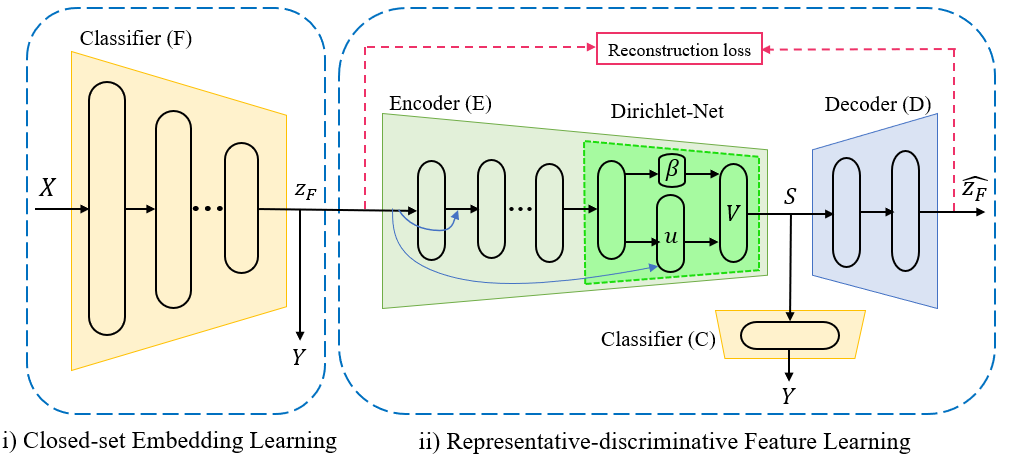}
\caption{An overview of the proposed framework: i) Closed-set embedding learning: the classifier {\it$F$} is trained on the spectral domain $X$ to produce latent discriminative embedding $\mathbf{z_F}$.  ii) Representative-discriminative feature learning: the encoder {\it$E$} takes the embedding feature $\mathbf{z_F}$ and derives the representative features $S$ using  a Dirichlet-Net. The classifier {\it$C$} applied on $S$ enhances the discriminative aspect of $S$, and the reconstruction error between the decoder output ($\mathbf{\hat{z}_F}$) and input to encoder ($\mathbf{z_F}$) enhances the representative aspect of $S$.
}
\label{fig:framework}
\end{figure}

One challenging issue of the open-set satellite land cover classification problem is that different classes may possess similar spectral characteristics. Thus, it is likely that an unknown class, whose spectral profile is close to that of a known class, may be misclassified as the known class. To address this issue, instead of detecting unknown classes on the image domain, we detect them on the embedding domain projected by a closed-set embedding layer, as shown in  Fig.~\ref{fig:framework}.i. The closed-set embedding layer increases the discriminative power of network to a large extent, such that unknown classes can be better recognized even if their spectra are similar to those of the known classes. The weights of the closed-set embedding layer are trained with a classifier $F$, which is further elaborated in Sec. \ref{sec:closedset}.

To recognize unknown classes in the embedding domain, 
we propose a multi-task representative-discriminative feature learning framework to boost both the representative and discriminative power of the extracted feature vector, such that it is more informative and effective to recognize unknown samples. This is shown in Fig.~\ref{fig:framework}.ii. The network consists of an encoder-decoder architecture with the  representative features $S$ extracted using a sparse Dirichlet encoder $E$, and a decoder formed by the bases shared among known classes. A classifier $C$ applied on $S$ is also included to further increase its discriminative capability. In this way, the data from unknown classes fed into the network would produce higher reconstruction error, thus can be detected accordingly. The details of network design are further elaborated in Sec. \ref{sec:discrepres}.

\subsection{Closed-set Embedding Learning}
\label{sec:closedset}
Given the set of sample pixels $X_k=\{\mathbf{x}_1, \mathbf{x}_2,\dots, \mathbf{x}_{N_k}\}$ from the known classes with each pixel, $\mathbf{x}_i$, being a high-dimensional vector recording the reflectance readings of different spectral bands in the hyperspectral image, the corresponding labels are denoted with $Y_k=\{y_1, y_2, \dots, y_{N_k}\}$, where $N_k$ is the number of known pixels and $\forall y_i \in \{1,2, \dots,L\}$, where $L$ is the number of known classes. To distinguish classes with similar spectral distributions, we project the input data $X_k$ from the image domain to the embedding domain $Z_F$. 
The projection is learned through a classifier, $F$, with parameters $\Theta_F$ and the embedded features, $\mathbf{z_F}$ is forced to be discriminative through the  cross-entropy loss,
\begin{equation}
  \mathcal{L}_f(\Theta_F) = -\frac{1}{N_k}\displaystyle\sum_{i=1}^{N_k}y_i \log[F(\mathbf{x}_i)],
\end{equation}
where $y_i$ is a one-hot encoded label and $F(\mathbf{x}_i)$ denotes the vector carrying the predicted probability score of the $i^{th}$ known sample. Such vector is generated by applying a softmax function on the features $\mathbf{z_F}$ in the embedding domain.

This general structure is sufficient for a common classification problem where the encountered classes are known. However, our goal is to increase the discriminative power of the features from classes with similar spectral characteristics. Therefore, we further increase the discriminative capacity of the  embedded features with the $l_1$-norm sparse constraint defined by

\begin{equation}
  \mathcal{L}_z(\Theta_F) = \frac{1}{N_k}\displaystyle\sum_{i=1}^{N_k}{\left\lVert \mathbf{z_F}_i \right\rVert},
\end{equation}
where $\mathbf{z_F}_i$ is the embedding feature vector learned by the classifier $F$. With such constraint, the embedded features from samples of different classes are more discriminative, even if their spectra are similar in the image domain.

\subsection{Multi-Task Representative-Discriminative Feature Learning}
\label{sec:discrepres}
With the proposed closed-set embedding layer, the samples are projected from the image domain to the embedding domain possessing more distinguishable features. In order to better identify unknown samples, both discriminative and representative nature of the samples need to be exploited. Previous approaches~\cite{oza2019deep, oza2019c2ae} usually train a general auto-encoder to reconstruct samples from known classes. When the samples from unknown classes are fed into the network, the reconstruction error is expected to be larger than that of the known classes in the ideal case, since the network weights are optimized during the training procedure using known samples. However, the challenge lies in the scenario when the unknown classes especially the ones close to the known classes may potentially contribute to small reconstruction error too which would lead to failure in detection.

In this work, instead of adopting a general-purpose auto-encoder, we propose a multi-task representative-discriminative feature learning framework to improve the detection accuracy. The purpose of this network is to decrease the reconstruction error of the known classes while increasing the reconstruction error of unknown classes intensively. 

Due to the resolution issue, each pixel in a satellite image usually covers a large geographical area or footprint (e.g., 30$\times$30m for Landsat-8), resulting in the so-called ``mixed pixel'' (i.e., each pixel tends to cover more than one constituent materials). These mixtures are generally assumed to be a linear combination of a few spectral bases with the corresponding mixing coefficients (or abundance). The proposed method is designed based on this assumption as shown in Eq.~\ref{equ:liner}. Assume that the feature vector of a sample from known classes $\mathbf{z_F}$ is a linear combination of a few bases $B$, and such bases are shared among features of the known classes. Thus, each sample of known classes can be decomposed with 
\begin{equation}
\mathbf{z_F} = \mathbf{s}B,
\label{equ:liner}
\end{equation}
where $\mathbf{s}$ denotes the proportional coefficients of the shared bases, which serves as a form of  ``representation'' of the embedding feature, that we refer to as the abundance.  The abundance vector, or representation, should satisfy two physical constraints, i.e., non-negative and sum-to-one. The samples from unknown classes are also able to be decomposed by Eq.~\ref{equ:liner} using shared bases of the known classes, $B$. However, since $B$ does not include the bases of the unknown classes, the distribution of its representations $\mathbf{s}$ should deviate from that of the known classes. Therefore, we design a network following the model of Eq.~\ref{equ:liner}, which  enforces $\mathbf{s}$ from known classes to follow a certain distribution. And if the network can extract $\mathbf{s}$ from unknown classes with similar distributions, then we expect they have high reconstruction errors.  

\begin{figure}[!t]
\centering
\includegraphics[scale=0.17]{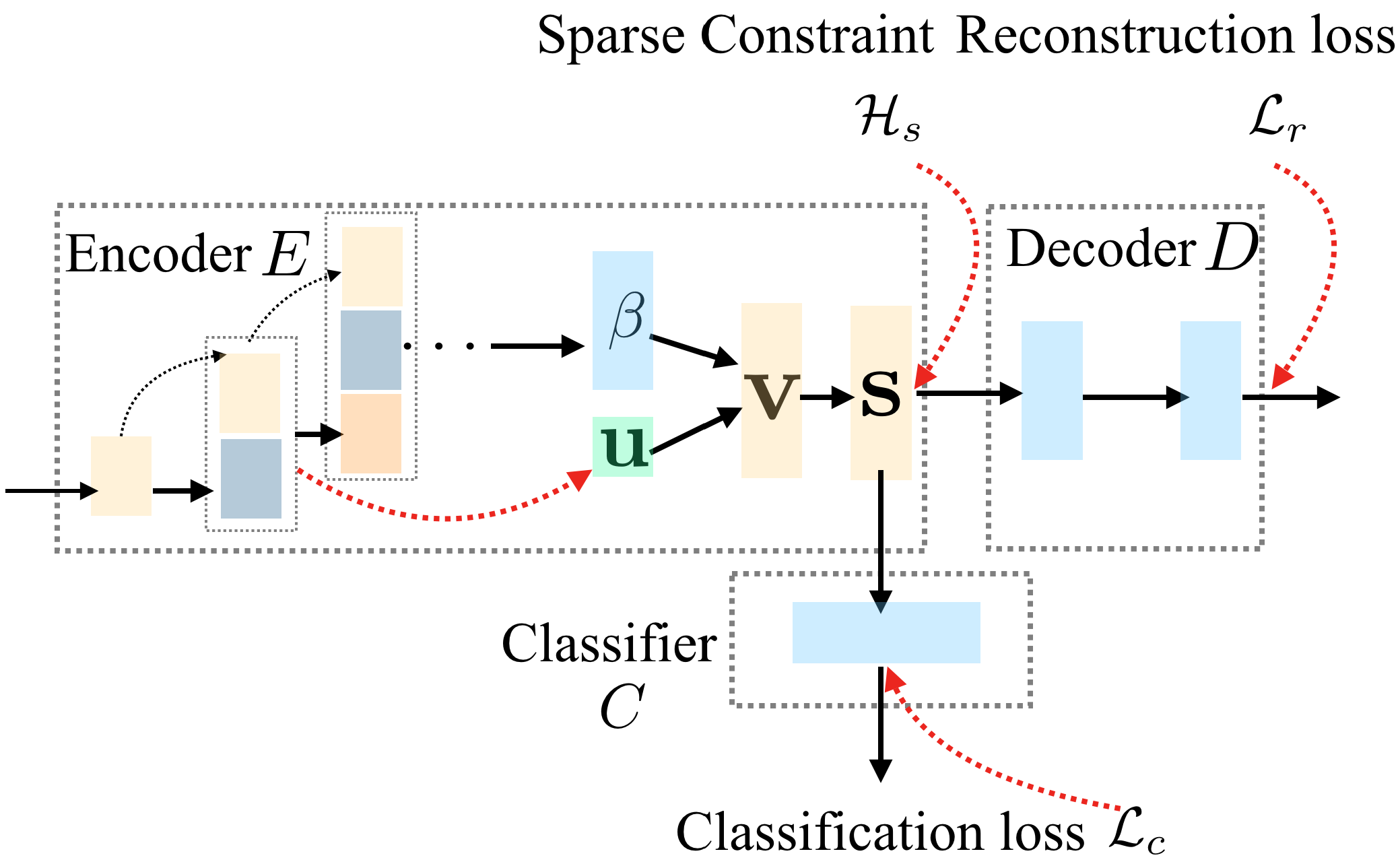}
\caption{The flowchart of the multi-task representative-discriminative feature learning framework.}
\label{fig:flow2}
\end{figure}
The flowchart of the proposed multi-task representative-discriminative feature learning is detailed in Fig.~\ref{fig:flow2}. The network performs both the reconstruction task and the classification task. The reconstruction branch consists of a sparse Dirichlet-based encoder $E$ with weights $\Theta_E$ and a decoder $D$ with weights $\Theta_D$.
The encoder and decoder can be defined by the functions $E: Z_F \to S$ and $D: S \to Z_F$, respectively, where $Z_F$ is the embedding space obtained by the closed-set classifier $F$, and $S$ is the abundance space of latent representations projected by the encoder $E$. The representations $\mathbf{s}$ in the latent space $S$ is enforced to follow a Dirichlet distribution.  And a sparse constraint is introduced to enhance the representativeness of $\mathbf{s}$. More details and justifications are provided below. 
In addition, $S$ is also enforced to be discriminative by the classifier $C$, which can be defined by the function $C: S \to Y$ with weights $\Theta_C$, where $Y$ is the space of known labels.

\subsubsection{Representative Feature Learning with Reconstruction}
\label{sec:dirichlet}
The reconstruction branch is constructed according to Eq.~\ref{equ:liner}, where the shared bases are embedded in the decoder $D$ of the network and the corresponding representation $\mathbf{s}$ is extracted with the encoder $E$. Since $\mathbf{s}$ denotes the proportional coefficients of the bases, we enforce it to follow a Dirichlet distribution meeting the non-negative and sum-to-one physical constraints. Following the work of~\cite{sethuraman1994constructive, nalisnick2017stick, qu2018unsupervised}, we adopt the stick-breaking structure in the encoder to enforce the representations $\mathbf{s}$ to follow the Dirichlet distribution.

In the stick-breaking structure, a single element $s_j$ in $\mathbf{s}$ can be expressed by
\begin{equation}
s_j =\left\{
\begin{array}{ll}
v_1 \quad & \text{for} \quad j = 1\\
v_j\prod_{o<j}(1-v_o) \quad &\text{for} \quad j>1 , 
\end{array}\right.
\label{equ:stick}
\end{equation}
where $v_j$ is drawn from a Kumaraswamy distribution, i.e., $v_j\sim \text{Kuma}(u, 1,\beta)$ as shown in Eq.~\eqref{equ:draw},
\begin{equation}
v_o\sim (1-(1-u^\frac{1}{\beta})).
\label{equ:draw}
\end{equation}
Then, there are two parameters used to extract representations $\mathbf{s}$, i.e., $u$ and $\beta$, both of which are hidden layers in the encoder of the network. 
A softplus activation function is adopted on the layer $\beta$ due to its non-negative property, and a sigmoid is used to map $u$ into the $(0,1)$ range at the layer $u$. More details of the stick-breaking structure can be found in~\cite{nalisnick2017stick} and~\cite{qu2018unsupervised}. 

In addition, the entropy function~\cite{huang2018sparse} is adopted to reinforce the sparsity of the representation layer. Let $\hat{s}_j = \frac{\vert s_j \vert}{\Vert \mathbf{s} \Vert}$, for each pixel, the entropy function is defined as,
\begin{equation}
\mathcal{H}_{s}(\Theta_E) = -\sum_{j=1}^{c}\hat{s}_j 
\log\hat{s}_j .
\label{equ:entropyfun}
\end{equation}
where $c$ is the dimension of the representation $\mathbf{s}$. 
The reconstruction loss $\mathcal{L}_r$ is adopted to reduce the reconstruction error of the known classes. It is defined by, 
\begin{align}
  \mathcal{L}_r(\{\Theta_E, \Theta_D\}) = \frac{1}{N_k}\displaystyle\sum_{i=1}^{N_k}{\left\lVert \mathbf{z_{F}}_i - \mathbf{\hat{z}_F}\right\rVert}_2,
\end{align}
where $\mathbf{z_{F}}_i$ is the embedding feature vector fed into the encoder $E$, and $\mathbf{\hat{z}_F}$ is the reconstructed $\mathbf{z_{F}}_i$ obtained from the decoder $D$.

\subsubsection{Discriminative Feature Learning with Classification Branch.} To further increase the discriminative capacity of the representations, a classifier is adopted on the representations $\mathbf{s}$ with the classification loss $\mathcal{L}_c$ defined as, 
\begin{align}
  \mathcal{L}_c(\{\Theta_E, \Theta_C\}) = \frac{1}{N_k}\displaystyle\sum_{i=1}^{N_k}y_i \log[E(\mathbf{z_{F}}_i)],
\end{align}
where $y_i$ is the ground truth label and $E(\mathbf{z_{F}}_i)$ denotes the representative feature vector of the $i^{th}$ known sample. Note that the weights of both the reconstruction branch and classifier $C$ are updated together, such that the learned representations can be both representative and discriminative. 

\subsection{Training Procedure and Network Settings}
\label{sec:training}
We first learn the embedding projection by optimizing the weights $\Theta_F$ of the classifier $F$ with the loss function,
\begin{align}
 \displaystyle{\min_{\Theta_F} \lambda_f\mathcal{L}_f + \lambda_z\mathcal{L}_z},
\end{align}
where $\lambda_f$ and $\lambda_z$ are two parameters to balance the trade-off between the cross-entropy loss and the sparsity loss. 

Then, having the learned embedding layer, the multi-task representative-discriminative feature learning network is trained to minimize both the reconstruction loss and the classification error of the known classes with loss function,
\begin{align}
 \displaystyle{\min_{\Theta_E, \Theta_D, \Theta_C} \lambda_r\mathcal{L}_r + \lambda_s\mathcal{H}_{s} + \lambda_c\mathcal{L}_c},
\end{align}
where $\lambda_r$, $\lambda_s$, and $\lambda_c$ are parameters balancing the trade-off between the reconstruction loss, the sparsity loss, and the classification loss. 

The structures of the four networks, $F$, $E$, $D$, and  $C$ are listed in Table \ref{tab:nodes}. 
\begin{table}[htb]
	\caption{The nodes in the proposed network}
	\label{tab:nodes}
	\begin{center}
		\begin{tabular}{c|cccc}
			\hline
			Networks&$F$&$E$&$D$&$C$\\
			\hline
			nodes &[512,1024,512,32,$L$]&[3,3,3,3, 10]&[10,10,$L$]&[$L$]\\
			\hline
		\end{tabular}
	\end{center}
\end{table}

\section{Experiments and Results}

In this section, the effectiveness of the proposed RDOSR method is evaluated on several widely used benchmark hyperspectral image datasets.
In addition, we demonstrate the generalization capacity of the proposed approach on RGB image datasets.
Furthermore, the contribution from each component of the proposed framework is analyzed through ablation study.

\subsection{Implementation Details}
We train the network, described in Sec. \ref{sec:network}, using an Adam optimizer \cite{kingma2014adam}, with a learning rate of $10^{-3}$. 
The classifier $F$ and the joint structure of encoder-decoder-classifier ($E$-$D$-$C$) are trained separately for a total number of $15$K epochs.
However, the other methods which do not have two separate components were trained for $6$K epochs.

For training the classifier $F$, $\lambda_f$ and $\lambda_z$ are set equal to $1$ and $0.1$, respectively. The weights for the reconstruction $\lambda_r$, sparsity $\lambda_s$, and classification $\lambda_c$ losses in training the $E$-$D$-$C$ structure
are set equal to $0.5$, $10^{-3}$, and $0.5$, respectively. The sparsity weight $\lambda_s$ is decayed with a weight decay of 0.9977. The classifier $F$ is trained until its accuracy reaches $0.9988$. It should be noted that all input data of the datasets are normalized to their mean value and unit variance. In addition, the feature vector obtained from the classifier $F$ is divided by 10 to avoid divergence.

One of the factors that affects the performance of the open-set recognition algorithm is Openness \cite{scheirer2012toward} of the problem, defined as,
\begin{align}
  Openness = 1 - \sqrt{\frac{2 \times N_{train}}{N_{test} + N_{target}}},
\end{align}
where $N_{train}$, $N_{test}$ and $N_{target}$ are the number of classes known during training, the number of classes given during testing, and the number of classes that need to be recognized correctly during testing phase, respectively. In the experiments, classes of each dataset is partitioned into known and unknown sets according to the Openness.

The code is written in TensorFlow, and all the experiments are performed on a desktop computer having GeForce GPU of 10 GB Memory.
The code is available at \textcolor{blue}{https://github.com/raziehkaviani/rdosr}.

\subsection{Metrics}

To compare performance of different methods, there are several metrics including overall accuracy or F-score on a combination of known and unknown classes, and Receiver Operating Characteristic (ROC). The first two metrics do not characterize the performance of the model well due to their sensitivity not only to the performance of the model in classifying the known classes, but also an arbitrary operating threshold for detecting unknown samples.

On the other hand, the ROC curve would illustrate the ability of a binary classification system (here, known vs. unknown detection) as a discrimination threshold is varied from the minimum to the maximum value of the given detection measure (here, reconstruction error).
Thus, it provides a measure free from calibration. To have a quantitative comparison, the area under the ROC (AUC) is computed in the experiments.

\subsection{Open-set Recognition for Hyperspectral Data}
The experiments are conducted on three hyperspectral image datasets:

\textbf{Pavia University (PU) and Pavia Center (PC).} Both PU and PC datasets were gathered over Northern Italy in 2011 by the Reflective Optics Systems Imaging Spectrometer which has a resolution of 1.3 m. The dimension of the PU dataset is $1096\times715$ pixels with 103 spectral bands, ranging from 430 to 860 nm. The PC dataset has $610\times340$ pixels with 102 spectral bands. The PU and PC datasets both include nine land cover categories.

\textbf{Indian Pines (IN).} The IN dataset was collected over Northwest Indiana in 1912 by Airborne Visible/Infrared Imaging Spectrometer (AVIRIS). It has a dimension of $145\times145$ with a resolution of 20 m by pixel and 200 spectral bands.
Its ground truth consists of 16 land cover classes.

We compare the performance of the proposed approach with three methods described in the following:

{\bf SoftMax}: In a neural network classifier, a common confidence-based approach to detect open-set examples is thresholding the SoftMax scores. We use network structure of the classifier $F$ without considering sparsity constraint.

{\bf OpenMax} \cite{bendale2016towards}: This approach calibrates the SoftMax scores in a classifier and augments them with a $N_k+1$ class for an unknown category. The replaced SoftMax layer with an OpenMax layer is used for open-set recognition. We adopt the classifier mentioned earlier in the SoftMax method, and use the Weibull fitting approach with parameter $Weibull\,tail\,size=10$ to generate OpenMax layer values.

{\bf AE+CLS}: Fully connected version of MLOSR \cite{oza2019deep} which utilizes a multi-task learning framework, composed of a classifier and a decoder with a shared feature extraction part, to detect open-set examples. To have a fair comparison with our approach, the encoder, decoder, and classifier are designed as our $E$ (without the Dirichlet-Net), $D$, $C$, and trained with $\mathcal{L}_r$ and $\mathcal{L}_c$ loss, with weights of 0.5. 

First, each of the $L$ classes is assumed to be unknown which equates to an openness of $2.99\%$, $2.99\%$, and $1.63\%$ for PU, PC, IN, respectively. The AUC values corresponding to choosing each of the $L$ labels as unknown are averaged and reported for each method in Table \ref{table:results_methods_datasets}. It can be observed that the proposed method outperforms other methods on all three datasets. See Supplement B for detailed comparison on the PU dataset.
The minor improvement on the PC dataset can be justified by its differentiated spectrum of different classes which diminishes the effect of the classifier $F$.

\begin{figure}[!t]
    \centering
    \subfloat[Open-set detection on PU dataset]{{\includegraphics[width=5.5cm]{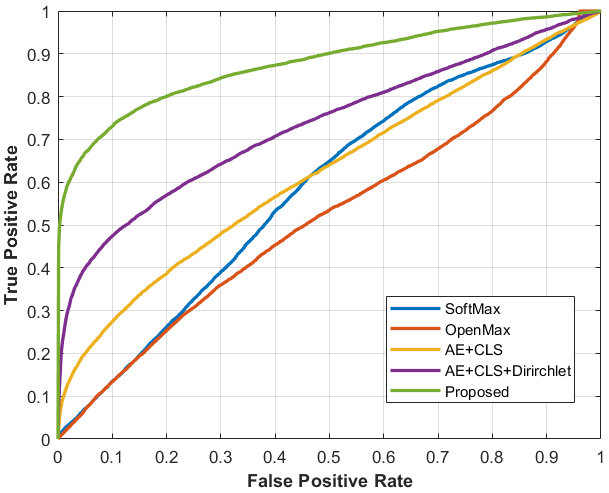} }}%
    \qquad
    \subfloat[Open-set detection on PC dataset]{{\includegraphics[width=5.5cm]{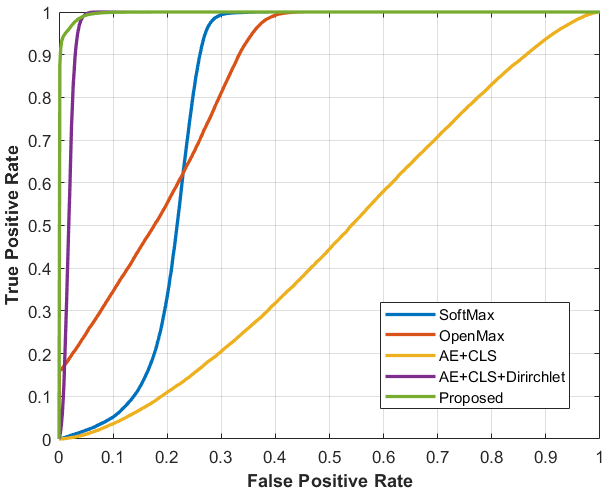} }}%
    \caption{Receiver Operating Curve curves for open-set recognition for PU and PC datasets, for $L=7$ (openness=$6.46\%$).}%
    \label{fig:ROC}%
\end{figure}

Second, for the Openness equal to $6.46\%$, the ROC curves of different methods for the PU and IN datasets are illustrated in Fig. \ref{fig:ROC}. As seen from the results on both datasets, the {\it AE+CLS+Dirichlet} method which adopts the Dirichlet net to the {\it AE+CLS} framework and the proposed method lie above all other methods. It should be noticed that our proposed method is able to detect unknown classes with $60\%$ and above $90\%$ accuracy and almost zero false detection for the PU and PC datasets, respectively.

\setlength{\tabcolsep}{4pt}
\begin{table}[!b]
\begin{center}
\caption{Area under the ROC curve for open-set detection. Results are averaged over $L$ partitioning of the selected dataset to $L-1$ known and $1$ unknown classes.}
\label{table:results_methods_datasets}
\begin{tabular}{lP{2cm}lP{1.5cm}lP{1.5cm}lP{1.5cm}l}
\hline\noalign{\smallskip}
\bf Method & \bf PU & \bf PC & \bf IN\\
\noalign{\smallskip}
\hline
\noalign{\smallskip}
SoftMax  & 0.385 & 0.816 & 0.555\\
OpenMax \cite{bendale2016towards} & 0.441 & 0.884 & 0.415\\
AE+CLS \cite{oza2019deep} & 0.586 & 0.757 & 0.669\\
AE+CLS+Dirichlet & 0.714 & 0.927 & 0.681\\
RDOSR (Ours) & {\bf 0.773} & {\bf 0.963} & {\bf 0.802}\\
\hline
\end{tabular}
\end{center}
\end{table}
\setlength{\tabcolsep}{1.4pt}

\begin{figure}[!t]
    \centering
    \subfloat[PU dataset]{{\includegraphics[width=5.5cm]{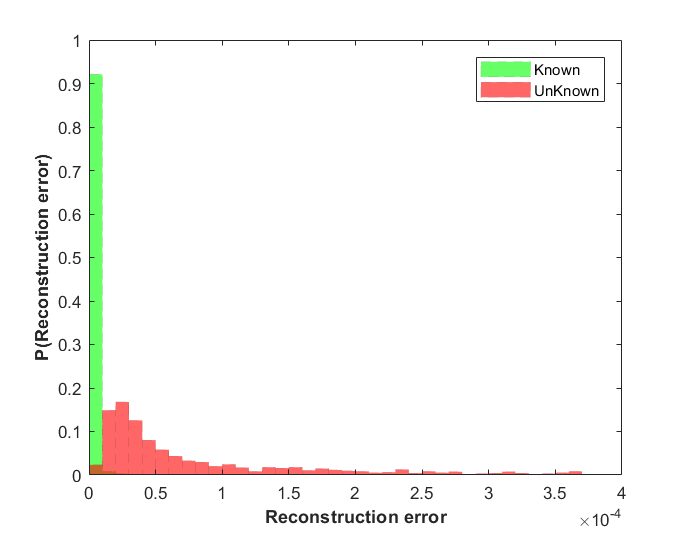} }}%
    \qquad
    \subfloat[PC dataset]{{\includegraphics[width=5.5cm]{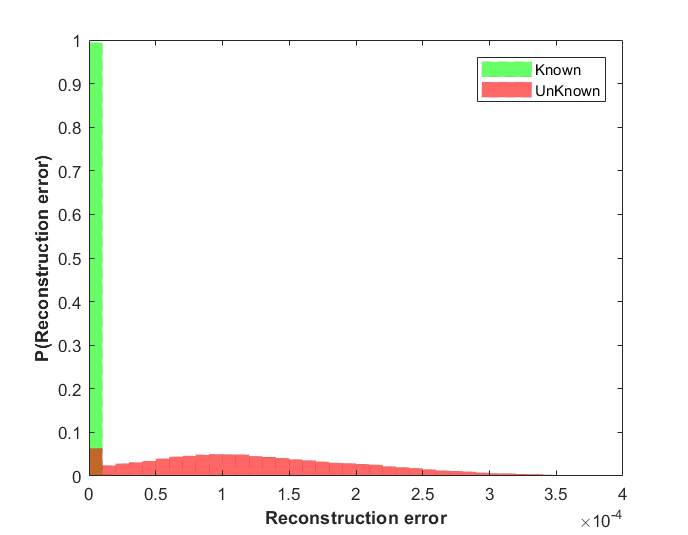} }}%
    \caption{Reconstruction error distribution of known and unknown classes using the proposed method for PU and PC datasets, for $L=8$.}%
    \label{fig:histogram}%
\end{figure}

Third, the histograms of reconstruction error for both known and unknown sets with Openness$=2.99\%$ are shown in Fig. \ref{fig:histogram}. It can be observed that the reconstruction errors corresponding to the known set have small values. However, the unknown set produces larger error due to mismatches in terms of representative and discriminative features learned from the known classes examples.

\subsection{Open-set Recognition for RGB Images}
To show the generalization capacity of the proposed method, we evaluate the performance of the proposed approach on two RGB datasets and compare with several state-of-the-art methods. For this purpose, the classifier $F$ performing pixel-wise classification is substituted with a DenseNet structure which takes a 2D image as input.

Following the protocol in \cite{neal2018open}, we sample 4 known classes from CIFAR10 \cite{krizhevsky2009learning} to have Openness=$13.39\%$ and 20 known classes out of 200 categories of TinyImageNet \cite{le2015tiny} resulting in an Openness of $57.35\%$.
Table \ref{table:results_methods_datasets_rgb} summarizes the results where the values other than the proposed RDOSR are taken from \cite{perera2020generative}. It can be observed that the proposed method has better performance over the compared methods, except for GDOSR \cite{perera2020generative}, on CIFAR10. However, it achieves significant improvement on TinyImageNet. It may be due to the similarity between the classes in TinyImageNet which hinders detecting unknown samples in the image space while RDOSR addresses this issue by operating in the embedding space.

\setlength{\tabcolsep}{4pt}
\begin{table}[!t]
\centering
\caption{Area under the ROC curve for Open-set recognition.}
\label{table:results_methods_datasets_rgb}
\begin{tabular}{lP{2.5cm}lP{2cm}lP{2cm}l}
\hline\noalign{\smallskip}
\bf Method & \bf CIFAR10 & \bf TinyImageNet\\
\noalign{\smallskip}
\hline
\noalign{\smallskip}
SoftMax  & 0.677 & \hfil0.577\\
OpenMax \cite{bendale2016towards} & 0.695 & \hfil0.576\\
OSRCI \cite{neal2018open} & 0.699 & \hfil0.586\\
C2AE \cite{oza2019c2ae} & 0.711 & \hfil0.581\\
GDOSR \cite{perera2020generative} & {\bf0.807} & \hfil0.608\\
RDOSR (Ours) & 0.744 & \hfil{\bf0.752}\\
\hline
\end{tabular}
\end{table}

\subsection{Ablation Study}
\begin{wrapfigure}{r}{0.6 \textwidth}
\centering
\includegraphics[scale=0.3]{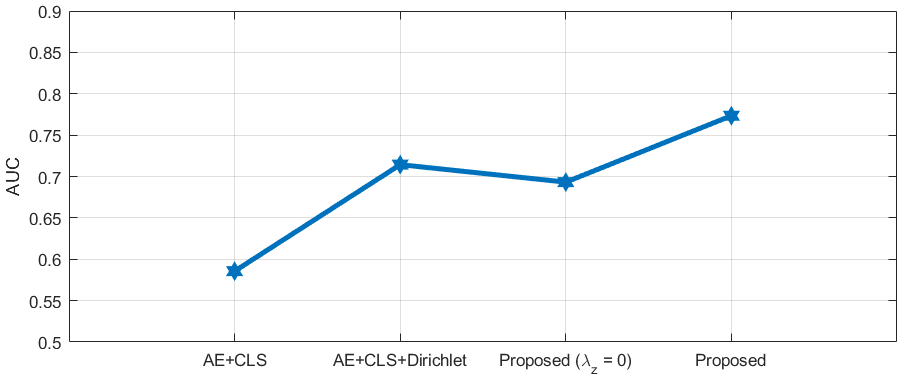}
\caption{Ablation study of the proposed method on PU dataset}
\label{fig:ablation_study}
\end{wrapfigure}
Starting with a baseline, {\it AE+CLS}, each component is gradually added to the framework to show its effectiveness. The results corresponding to the ablation study are shown in Fig. \ref{fig:ablation_study}. It can be seen that employing the baseline structure applied on the spectra domain has the worst performance. However, adding the Dirichlet-based network makes a major improvement due to applying physical constraints on the latent space learned by the encoder $E$. Directly performing open-set recognition on an embedding space causes instability problem which is confirmed by a performance drop compared to the {\it AE+CLS+Dirichlet} method. Our proposed method addresses the instability issue by adopting a sparsity constraint on the embedding feature vector $\mathbf{z_F}$. As seen from Fig. \ref{fig:ablation_study}, our proposed method achieves the highest AUC value compared to three other baseline methods.

\section{Conclusions}
We studied the challenging problem of open-set land cover recognition in satellite images. Although inherently a classification problem, both representative and discriminative features need to be learned in order to best characterize the difference between known and unknown classes. We presented the transformation among three spaces, that is, the original image space, the embedding feature space, and the abundance space, where features with both representative and discriminative capacity can be learned to maximize success rate. The proposed multi-tasking representative-discriminative learning structure was evaluated on three hyperspectral and two RGB image datasets and exhibited significant improvement over state-of-the-art open-set recognition algorithms.

\clearpage
%
%
\bibliographystyle{splncs}
\bibliography{Razieh_eccv20_}

\clearpage

\begin{center}
\textbf{\Large Supplementary Material: Representative-Discriminative Learning for Open-set Land Cover Classification of Satellite Imagery}
\end{center}

\setcounter{equation}{0}
\setcounter{figure}{0}
\setcounter{table}{0}
\makeatletter
\renewcommand{\theequation}{S\arabic{equation}}
\renewcommand{\thefigure}{S\arabic{figure}}
\renewcommand{\thetable}{S\arabic{table}}

\section*{Supplementary Contents}

Section A presents a visual comparison of data used for open-set recognition among the three spaces, i.e., the image space ($X$), the embedding space ($Z_f$), and the abundance space ($S$). Section B shows the performance of open-set recognition if conducted on the image space ($X$) or the embedded space ($Z_f$).

\section*{A \hspace{0.5cm} A visual comparison among spaces $X$, $Z_f$, and $S$}

\begin{figure}
\centering
\includegraphics[scale=0.5]{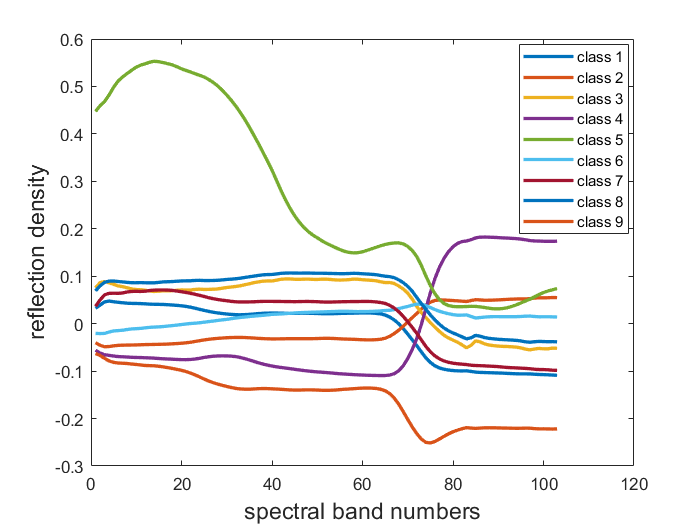}
\caption{Mean of samples in space $X$, belonging to classes 1 to 9, using the PU dataset}
\label{fig:spaceX}
\end{figure}

\begin{figure}%
    \centering
    \subfloat{{\includegraphics[scale=0.213]{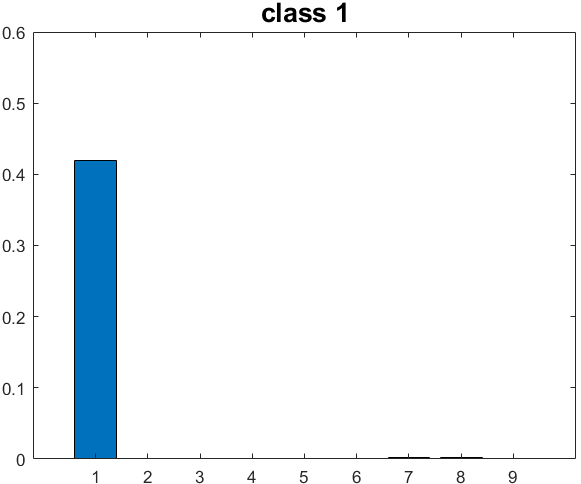}}}%
    \qquad
    \subfloat{{\includegraphics[scale=0.213]{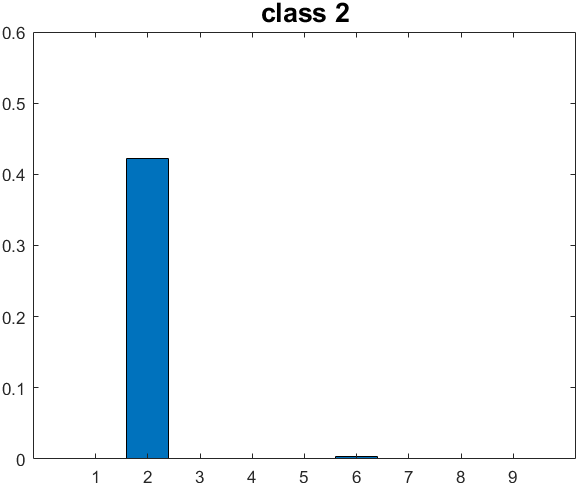} }}%
    \qquad
    \subfloat{{\includegraphics[scale=0.213]{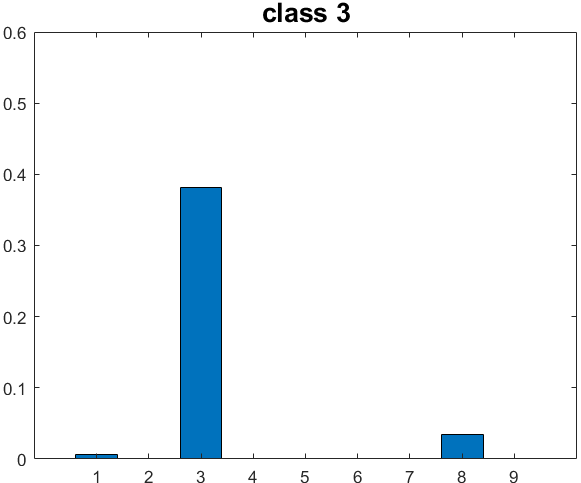} }}%
    \qquad
    \subfloat{{\includegraphics[scale=0.213]{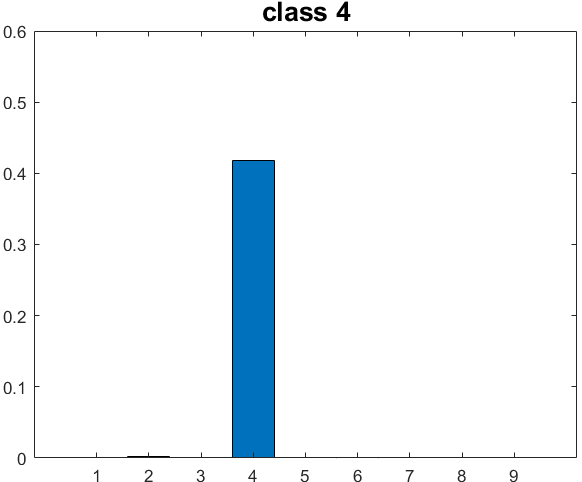} }}%
    \qquad
    \subfloat{{\includegraphics[scale=0.213]{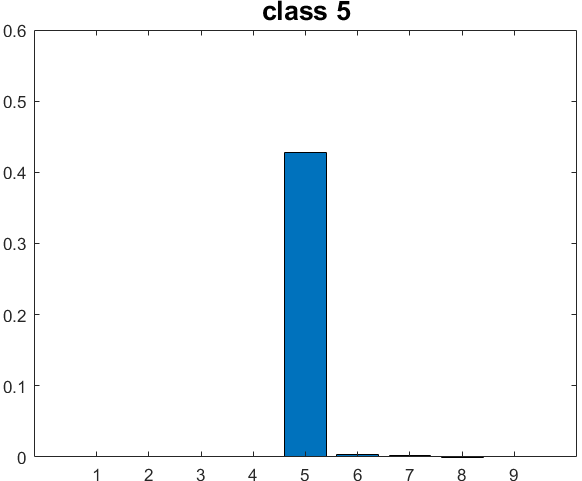} }}%
    \qquad
    \subfloat{{\includegraphics[scale=0.213]{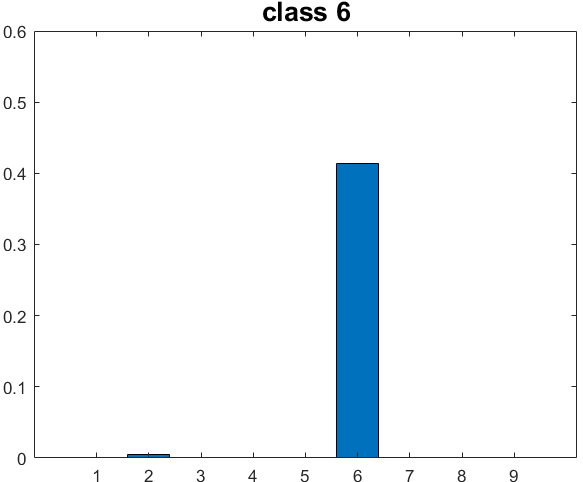} }}%
    \qquad
    \subfloat{{\includegraphics[scale=0.213]{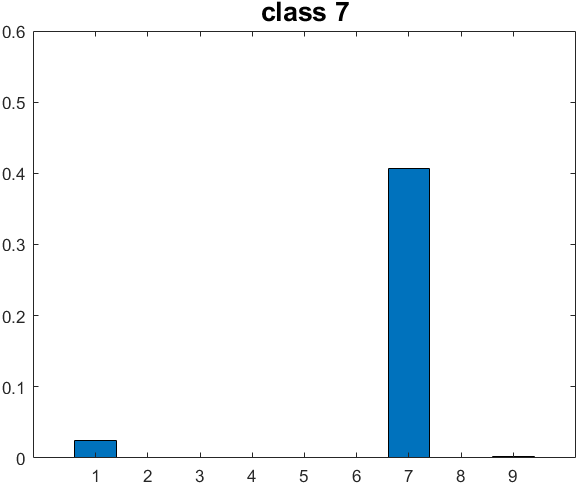} }}%
    \qquad
    \subfloat{{\includegraphics[scale=0.213]{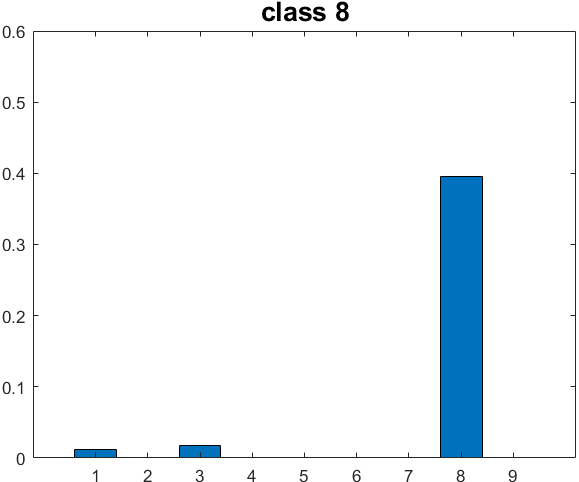} }}%
    \qquad
    \subfloat{{\includegraphics[scale=0.213]{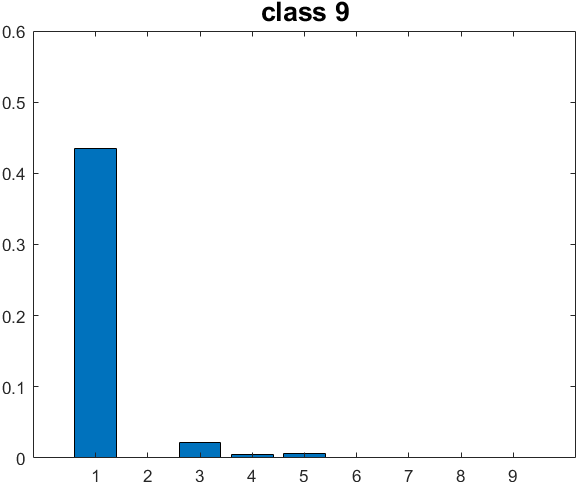} }}%
    \caption{Mean of samples in space $Z_f$, belonging to classes 1 to 9, using the PU dataset}%
    \label{fig:spaceZf}%
\end{figure}

\begin{figure}[t]
    \centering
    \subfloat{{\includegraphics[scale=0.213]{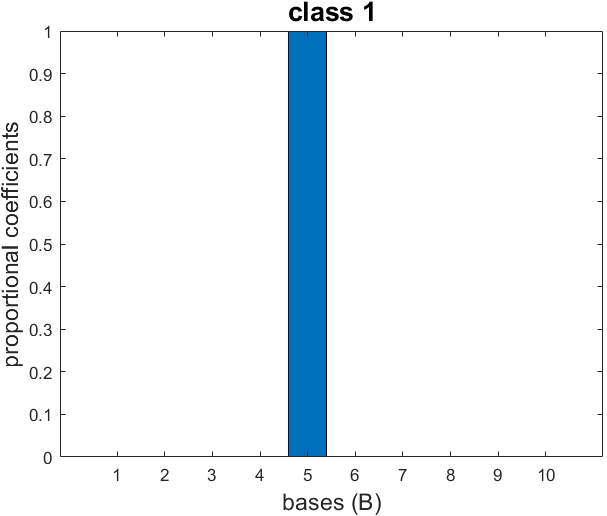}}}%
    \qquad
    \subfloat{{\includegraphics[scale=0.213]{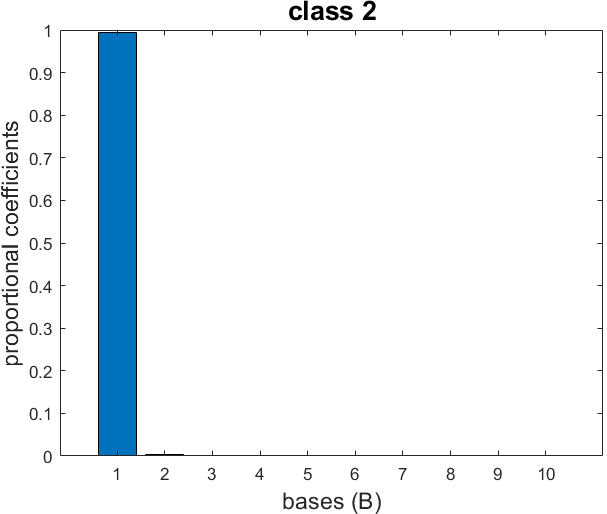} }}%
    \qquad
    \subfloat{{\includegraphics[scale=0.213]{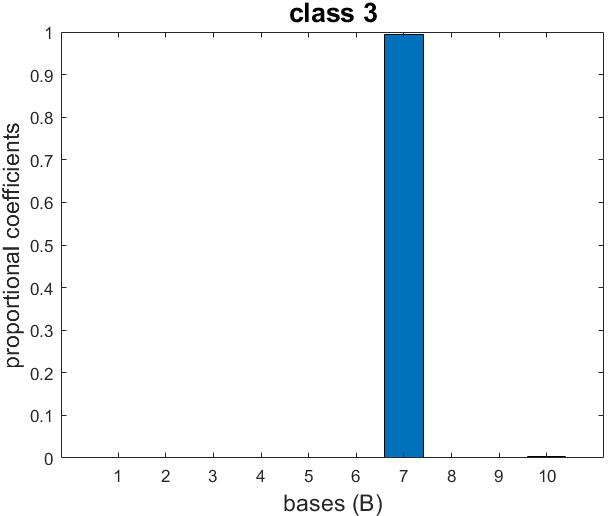} }}%
    \qquad
    \subfloat{{\includegraphics[scale=0.213]{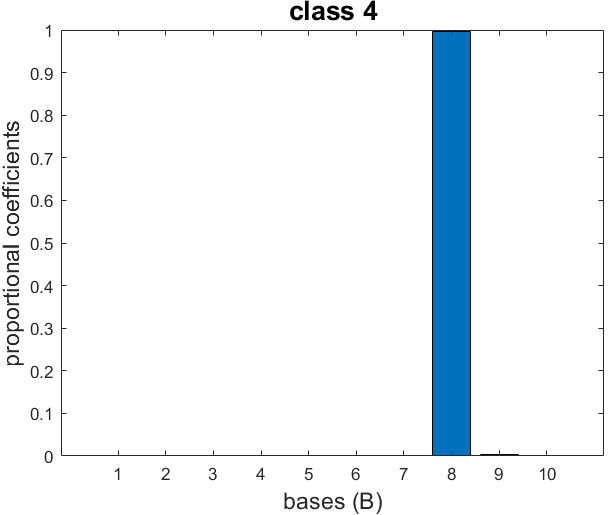} }}%
    \qquad
    \subfloat{{\includegraphics[scale=0.213]{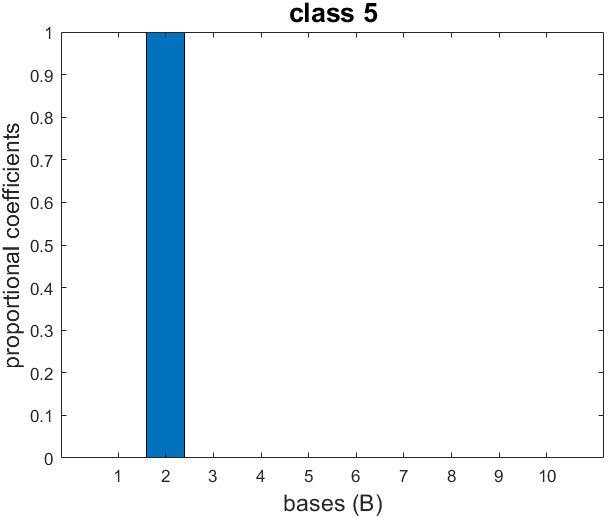} }}%
    \qquad
    \subfloat{{\includegraphics[scale=0.213]{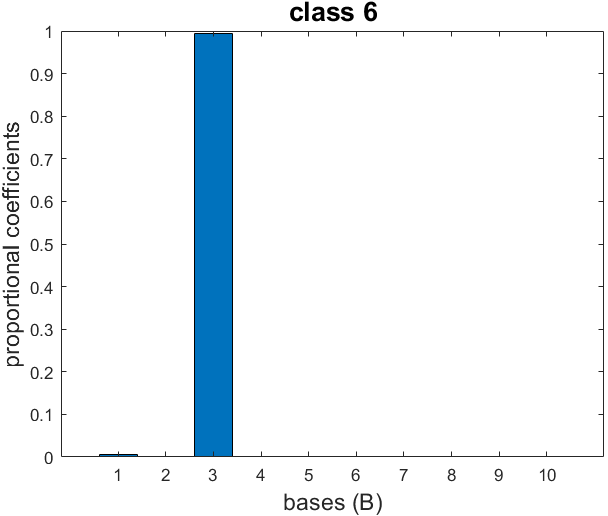} }}%
    \qquad
    \subfloat{{\includegraphics[scale=0.213]{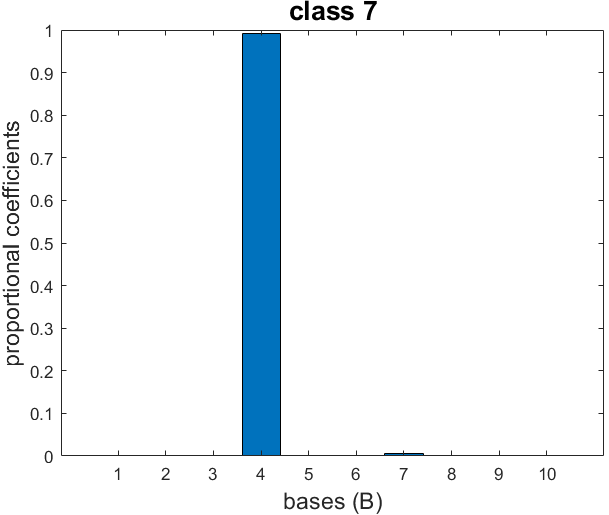} }}%
    \qquad
    \subfloat{{\includegraphics[scale=0.213]{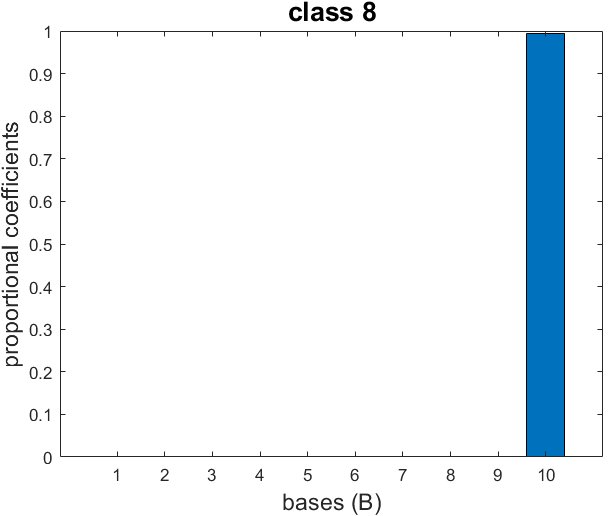} }}%
    \qquad
    \subfloat{{\includegraphics[scale=0.213]{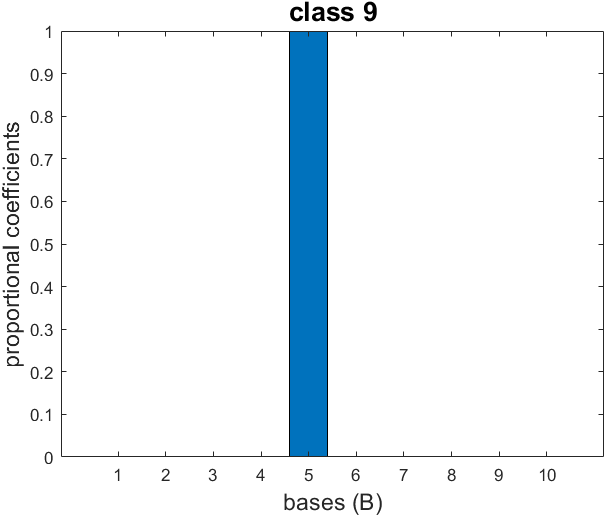} }}%
    \caption{Mean of samples in space $S$, belonging to classes 1 to 9, using the PU dataset}%
    \label{fig:spaceS}%
\end{figure}

Figures \ref{fig:spaceX}, \ref{fig:spaceZf}, and \ref{fig:spaceS} illustrate the mean of samples belonging to different classes in spaces $X$, $Z_f$, and $S$, respectively, using the PU dataset. The feature vectors learned through $F$ and $E$, shown in Figs. \ref{fig:spaceZf} and \ref{fig:spaceS}, respectively, are sparse due to the sparsity constraint. 

It can be seen from Fig. \ref{fig:spaceX} that the spectrum of samples belonging to class 3 and 8 are close. However, the feature vectors, learned through $F$, corresponding to class 3 and 8 are more discriminative. Further, the discriminative and representative characteristics of the features are enhanced through encoder $E$, as illustrated in Fig. \ref{fig:spaceS}.
\\

\section*{B \hspace{0.5cm} Comparison of open-set recognition performing on space $X$ and $Z_f$}

To better compare the effectiveness of performing open-set recognition in spaces $X$ and $Z_f$, we show the results of performing in each space separately using the PU dataset in Table \ref{table:results_spaces}. 

Comparing the results of $AE+CLS$ and $AE+CLS+Dirichlet$ approaches performed on spaces $X$ and $Z_f$, it can be seen that discriminative features learned through the classifier $F$ contribute to a substantial improvement. In addition, the Dirichlet network plays more critical role when performing open-set recognition in space $X$ as compared to space $Z_f$.

\setlength{\tabcolsep}{3pt}
\begin{table}[!t]
\begin{center}
\caption{Area under the ROC curve for open-set recognition. Results are from partitioning PU dataset to $L-1$ known and the mentioned unknown classes, (openness=$2.99\%$). Note that $L$ denotes the number of classes in the original PU dataset}
\label{table:results_spaces}
\begin{tabular}{l|c|llllllllll}
\hline\noalign{\smallskip}
\bf Space & \bf Method & \bf 1 & \bf 2 & \bf 3 & \bf 4 & \bf 5 & \bf 6 & \bf 7 & \bf 8 & \bf 9 & \bf Avg.\\
\noalign{\smallskip}
\hline
\noalign{\smallskip}
\multirow{4}{*}{$X$} & SoftMax  & 0.54 & 0.52 & 0.51 & 0.42 & 0.14 & 0.38 & 0.23 & 0.64 & 0.09 & 0.39\\
& OpenMax [26] & 0.67 & 0.37 & 0.45 & 0.40 & 0.99 & 0.35 & 0.12 & 0.57 & 0.04 & 0.44\\ 
& AE+CLS [30] & 0.51 & 0.53 & 0.54 & 0.83 & 1.0 & 0.46 & 0.48 & 0.46 & 0.46 & 0.59\\ 
& AE+CLS+Dirichlet & 0.79 & 0.69 & 0.47 & 0.90 & 1.0 & 0.64 & 0.47 & 0.48 & 0.97 & 0.71\\
\hline
\multirow{2}{*}{$Z_f$} & AE+CLS & 0.91 & 0.70 & 0.68 & 0.72 & 1.0 & 0.62 & 0.46 & 0.66 & 0.94 & 0.74\\
& AE+CLS+Dirichlet & 0.91 & 0.70 & 0.71 & 0.72 & 1.0 & 0.68 & 0.51 & 0.80 & 0.93 & {\bf 0.77}\\
\hline
\end{tabular}
\end{center}
\end{table}
\setlength{\tabcolsep}{1.4pt}

\end{document}